\documentclass[lettersize,journal]{IEEEtran}

\usepackage[T1]{fontenc}
\usepackage[utf8]{inputenc}

\usepackage{amsmath,amsfonts}
\usepackage{algorithmic}
\usepackage{algorithm}
\usepackage{array}
\usepackage[caption=false,font=normalsize,labelfont=sf,textfont=sf]{subfig}
\usepackage{textcomp}
\usepackage{url}
\usepackage{cite}
\usepackage{verbatim}
\usepackage{graphicx}
\usepackage{cite}
\usepackage{multirow}
\usepackage{booktabs}
\usepackage[hidelinks]{hyperref}
\DeclareRobustCommand{\textproc}[1]{\textsc{#1}}

\usepackage{siunitx}
\usepackage[acronym]{glossaries}
\makenoidxglossaries
\setacronymstyle{long-short}

\DeclareUnicodeCharacter{2212}{\ensuremath{-}} 
\DeclareUnicodeCharacter{2013}{--}             
\DeclareUnicodeCharacter{2014}{---}            

\newacronym{ai}{AI}{Artificial Intelligence}
\newacronym{artex}{ARTEX}{Army Technological Exercise}
\newacronym{astar}{A*}{A-star}
\newacronym{ba}{BA*}{Bidirectional A*}
\newacronym{bfs}{BFS}{Breadth-First Search}
\newacronym{c2}{C2}{Command and Control}
\newacronym{c4isr}{C4ISR}{Command, Control, Communications, Computers, Intelligence, Surveillance and Reconnaissance}
\newacronym{cos}{COS}{Carta de Uso e Ocupação do Solo}
\newacronym{cpa}{CPA}{Closest Point of Approach}
\newacronym{dem}{DEM}{Digital Elevation Model}
\newacronym{dijkstra}{Dijkstra's}{Dijkstra's Algorithm}
\newacronym{dimacs}{DIMACS}{Center for Discrete Mathematics and Theoretical Computer Science}
\newacronym{dstar}{D*}{D-star}
\newacronym{gis}{GIS}{Geographic Information System}
\newacronym{gnss}{GNSS}{Global Navigation Satellite System}
\newacronym{gsm}{GSM}{Global System for Mobile Communications}
\newacronym{isr}{ISR}{Intelligence, Surveillance, and Reconnaissance}
\newacronym{kpi}{KPI}{Key Performance Indicator}
\newacronym{lpa}{LPA*}{Lifelong Planning A*}
\newacronym{nato}{NATO}{North Atlantic Treaty Organization}
\newacronym{rcspp}{RCSPP}{Resource-Constrained Shortest Path Problem}
\newacronym{repmus}{REPMUS}{Robotic Experimentation and Prototyping using Maritime Unmanned Systems}
\newacronym{snr}{SNR}{Signal-to-Noise Ratio}
\newacronym{ugv}{UGV}{Unmanned Ground Vehicle}
\newacronym{wc-a}{WC-A*}{Weight--Constrained A*}
\newacronym{wc-ba}{WC-BA*}{Weight--Constrained Bidirectional A*}
\newacronym{wc-ebba}{WC-EBBA*}{Weight--Constrained Enhanced Bidirectional Bounded A*}
\newacronym{argus}{ARGUS}{Adaptive Route Guidance and Update System}
\newacronym{apulse}{APULSE}{A*-Pulse}

\begin{document}

\title{APULSE: A Scalable Hybrid Algorithm for the RCSPP on Large-Scale Dense Graphs}

\author{
    \begin{tabular}[t]{c} 
        Nuno Alexandre Duarte Soares \\
        \textit{Academia Militar} \\
        Lisboa, Portugal \\
        soares.nad@academiamilitar.pt
    \end{tabular}
    \hspace{2cm} 
    \begin{tabular}[t]{c} 
        António Manuel Raminhos Cordeiro Grilo \\
        \textit{INESC INOV} \\
        \textit{Instituto Superior Técnico (IST)} \\
        \textit{Universidade de Lisboa} \\
        Lisboa, Portugal \\
        antonio.grilo@inov.pt
    \end{tabular}
}

\maketitle

\begin{abstract}
The resource-constrained shortest path problem (RCSPP) is a fundamental NP-hard optimization challenge with broad applications, from network routing to autonomous navigation. This problem involves finding a path that minimizes a primary cost subject to a budget on a secondary resource. While various RCSPP solvers exist, they often face critical scalability limitations when applied to the large, dense graphs characteristic of complex, real-world scenarios, making them impractical for time-critical planning. This challenge is particularly acute in domains like mission planning for unmanned ground vehicles (UGVs), which demand solutions on large-scale terrain graphs. This paper introduces APULSE, a hybrid label-setting algorithm designed to efficiently solve the RCSPP on such challenging graphs. APULSE integrates a best-first search guided by an A* heuristic with aggressive, Pulse-style pruning mechanisms and a time-bucketing strategy for effective state-space reduction. A computational study, using a large-scale UGV planning scenario, benchmarks APULSE against state-of-the-art algorithms. The results demonstrate that APULSE consistently finds near-optimal solutions while being orders of magnitude faster and more robust, particularly on large problem instances where competing methods fail. This superior scalability establishes APULSE as an effective solution for RCSPP in complex, large-scale environments, enabling capabilities such as interactive decision support and dynamic replanning.
\end{abstract}


\section{Introduction} \label{sec:introduction} \IEEEPARstart{T}{he} \gls{rcspp} is a fundamental NP-hard optimization challenge with broad applications, from network routing and logistics to autonomous navigation \cite{Lozano2013}. The problem involves finding a path that minimizes a primary cost (e.g., risk, energy, distance) subject to a budget on a secondary resource (e.g., time, battery life, monetary cost). While the \gls{rcspp} is well-studied, a critical gap persists between the vast body of theoretical solutions and the demands of practical, time-critical applications. The core challenge lies in algorithmic scalability. Many real-world problems, such as robotic pathfinding on high-resolution terrain grids or routing in complex topologies, must be modeled as large-scale and dense graphs. On such graphs, existing \gls{rcspp} solvers, which are typically based on label-setting algorithms, suffer from an exponential proliferation of non-dominated labels. This leads to prohibitive runtimes, rendering them impractical for any scenario requiring rapid decision-making \cite{Pugliese2013}.

This scalability bottleneck is not merely an academic limitation; it directly prevents the implementation of critical capabilities across numerous domains. The need for algorithms that are simultaneously near-optimal, lightweight, and fast is paramount. An algorithm that requires minutes or hours for a single query is relegated to offline, static planning. In contrast, a high-performance solver that finds high-quality solutions in seconds unlocks transformative capabilities. These include interactive decision support, where a user can perform rapid "what-if" analysis by varying constraints (e.g., "how much safer is the path if I allow 10 more minutes?"), and dynamic replanning, the ability for an autonomous system to compute a new, optimal path in real-time in response to new intelligence or a changing environment \cite{koenig2004lifelong}.

The state-of-the-art in solving the \gls{rcspp} has converged on sophisticated label-setting algorithms that combine heuristic guidance (A*), bidirectional search, and dominance-based pruning \cite{Ahmadi2021}\cite{Thomas2018}. While these methods have shown success on sparse networks (e.g., road graphs), they consistently exhibit performance bottlenecks on the dense, grid-based graphs that characterize many relevant problems. The overhead associated with extensive label management and dominance checks in these high-connectivity environments remains the primary unsolved obstacle. This creates a critical gap for a solver that is explicitly designed for scalability on these challenging graph structures, even if it must trade absolute exactness for operational speed.

To address this scalability challenge, this paper introduces APULSE, a hybrid search algorithm specifically designed to find near-optimal paths on large-scale, dense \gls{rcspp} instances with high performance. APULSE synergistically integrates a best-first search strategy guided by an A* heuristic with aggressive, Pulse-style pruning mechanisms \cite{Lozano2013} and an efficient time-bucketing technique for state-space reduction. This hybrid design is not tailored to a specific cost model; it is engineered to exploit the structure of dense graphs to drastically reduce the search space while maintaining high solution quality.

The efficacy of APULSE is demonstrated through a computational study against state-of-the-art exact \gls{rcspp} solvers using a large-scale and dense graph derived from a real-world \gls{ugv} planning scenario, which serves as a challenging and representative testbed for this class of problem. The results demonstrate that APULSE consistently finds near-optimal solutions while being orders of magnitude faster and more robust, successfully solving 100\% of tested instances where competing methods fail or exceed operational time limits. This superior scalability establishes APULSE as an effective solution for the \gls{rcspp} in complex, large-scale environments.

The main contributions of this paper are the following:  
\begin{itemize}  
\item The introduction of APULSE, a hybrid algorithm whose design—integrating A* guidance, Pulse-style pruning, and time-bucketing—is specifically tailored to overcome the scalability bottlenecks of existing solvers on large-scale, dense graphs.  
\item An empirical demonstration in the context of autonomous navigation, where APULSE is benchmarked against state-of-the-art solvers on large-scale terrain graphs representative of military environments. The results show that a unidirectional search approach, when synergistically combined with aggressive pruning and state-space reduction mechanisms, achieves superior scalability and performance compared to more complex bidirectional algorithms. A key contribution of this study is the conclusion that, on large-scale dense graphs where bidirectional methods fail, a well-designed unidirectional solver provides a more robust and efficient solution. 
\end{itemize}

The remainder of this paper is organized as follows. Section~\ref{sec:related_work} reviews the relevant literature on \gls{rcspp} algorithms. Section~\ref{sec:problem_formulation} presents the mathematical formulation of the \gls{rcspp} as applied to the benchmark problem. Section~\ref{sec:apulse_algorithm} details the proposed APULSE algorithm. Section~\ref{sec:computational_study} presents the computational experiments and benchmark results. Section~\ref{sec:discussion} discusses the implications of the results, and Section~\ref{sec:conclusion} concludes the paper and suggests directions for future research.

\section{Related Work}
\label{sec:related_work}
The problem of finding a path in a graph subject to one or more constraints is a cornerstone of operations research and computer science. The specific formulation addressed in this paper, the \gls{rcspp}, seeks to find a path that minimizes a primary cost metric while adhering to a budget on a secondary resource \cite{Irnich2005}. The \gls{rcspp} is NP-hard, a characteristic that precludes the existence of polynomial-time algorithms for its exact solution and has motivated extensive research into efficient, practical solvers \cite{Pugliese2013}.

Early and foundational approaches to the \gls{rcspp} are based on dynamic programming, typically implemented as label-correcting or label-setting algorithms \cite{Lozano2013}. These methods explore the state space by propagating labels, where each label represents a partial path to a node, storing its accumulated cost and resource consumption. The core pruning mechanism in these algorithms is the concept of dominance: a label (path) is discarded if another label reaches the same node with an equal or lower cost in all dimensions (e.g., lower risk and lower or equal time). While theoretically sound, these classical methods suffer from a critical scalability issue: the number of non-dominated labels at each node can grow exponentially with the size of the graph, leading to prohibitive memory usage and computation times on large networks \cite{Pugliese2013}.

To overcome these limitations, modern research has focused on enhancing label-setting algorithms with techniques that more effectively manage the search space. A dominant trend has been the integration of heuristic guidance, most notably through the A* algorithm and its variants \cite{Hart1968}. A* prioritizes the expansion of nodes that are not only cheap to reach from the start but are also estimated to be close to the goal, focusing the search on the most promising regions of the graph. Bidirectional search strategies, which explore simultaneously from the start and goal nodes, have also been employed to reduce the overall search space \cite{Thomas2018}\cite{Holte2016}.

A parallel line of research has developed aggressive, real-time pruning techniques. The family of Pulse algorithms, which are essentially highly optimized depth-first searches, excel at this \cite{Cabrera2020}. They employ strong feasibility and optimality cuts based on incumbent solutions and resource consumption to discard vast portions of the search tree dynamically. While pure Pulse algorithms lack the goal-oriented guidance of \gls{astar}, their powerful pruning capabilities are a critical component of state-of-the-art hybrid solvers.

Finally, for very large networks, state-space reduction techniques are often essential. A prevalent method is the discretization of resource dimensions into "buckets," where labels are grouped into fixed-width intervals \cite{Irnich2005}\cite{Delling2009}. By relaxing the dominance rule to apply only within the same bucket, this technique dramatically reduces the number of labels that need to be stored and processed, trading a controlled degree of optimality for significant gains in performance. Recent benchmarks of \gls{rcspp} solvers on large-scale road networks have shown that hybrid algorithms combining these three pillars—A* guidance, advanced pruning, and state-space reduction—consistently achieve the best performance \cite{Ahmadi2024EnhWCSP}\cite{Ahmadi2021}.

However, a significant gap remains for solving the \gls{rcspp} on dense, grid-based graphs. These structures, which are characteristic of many real-world problems such as terrain-based path planning for \glspl{ugv}, pose unique challenges. The high connectivity and uniformity often found in these graphs can diminish the effectiveness of standard bidirectional heuristics and exacerbate the problem of label proliferation. This context demands an algorithm specifically tailored to balance aggressive pruning with effective heuristic guidance. The proposed algorithm, APULSE, is designed to fill this gap by creating a synergistic combination of A*, Pulse-style pruning, and time-bucketing, optimized for scalability on such challenging graph structures.

\section{Problem Formulation}
\label{sec:problem_formulation}

With no loss of generality, the present work was motivated by the \gls{ugv} path planning problem in military missions, particularly in scenarios of electronic warfare and contested environments. In such contexts, an autonomous \gls{ugv} must traverse a hostile area while minimizing its probability of detection and destruction by adversarial sensors or weapon systems. The \gls{ugv} path planning task is therefore modeled as the search for an optimal path through a weighted, directed graph $G = (V, E)$, where both mission efficiency (time) and survivability (risk) are taken into account.

This graph provides a discrete abstraction of a continuous operational environment, typically characterized by heterogeneous terrain and varying threat exposure. The resulting graph is large-scale and dense, reflecting the fine spatial resolution required for tactical navigation. Two distinct cost metrics are encoded: a \emph{resource cost} (traversal time) associated with edges and an \emph{optimization cost} (risk) associated with vertices. The joint minimization of these costs defines the multi-objective nature of the problem.

\subsection{Graph Representation and Resource Cost (Time)}
\begin{figure}[t]
    \centering
    \includegraphics[width=0.9\columnwidth]{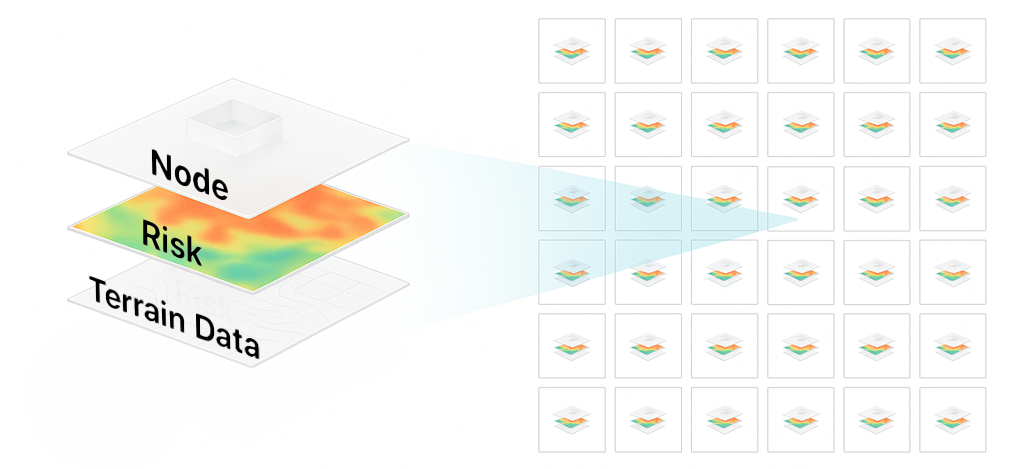}
    \caption{Conceptual illustration of the data layers associated with each node in the graph. Each node encapsulates geospatial attributes (terrain type, slope) and a computed risk value, forming the basis for the multi-cost pathfinding problem.}
    \label{fig:node_layers}
\end{figure}

The environment is discretized into a regular grid, where the set of vertices $V$ corresponds to cell centroids and the set of edges $E$ represents feasible (e.g., 8-connectivity) movements between adjacent cells. Each node $v \in V$ is enriched with geospatial data such as terrain type, slope, and vegetation density \cite{cos_dgt}, as illustrated in Fig.~\ref{fig:node_layers}. These environmental factors directly influence the vehicle’s traversability and, consequently, its estimated speed.

Beyond the raw environmental layers, a terrain-dependent mobility model is used to estimate the \emph{maximum feasible speed} for the UGV. Based on the combination of terrain type (e.g., pavement, scrubland, forest) and the local slope computed from the DEM, a velocity matrix is constructed for the specific \gls{ugv} model. This matrix encodes, for each terrain--slope pair, the expected maximum speed that the UGV can safely and realistically achieve. As a result, it serves as the operational definition of $v_{\text{terrain}}$ in the cost formulation.

Edges crossing areas where the mobility model yields higher speeds are therefore favoured, while edges traversing dense vegetation or steep gradients incur lower $v_{\text{terrain}}$ values. These slower segments increase traversal time and, by extension, degrade survivability: the longer the UGV remains in a given region, the higher the cumulative exposure to hostile fires or detection systems. Consequently, terrain-induced speed limitations directly translate into increased resource consumption within the RCSPP formulation.

The resource cost of each edge $(u, v) \in E$ is the traversal time,
\begin{equation}
t(u,v) = \frac{d(u,v)}{v_{\text{terrain}}(u,v)},
\label{eq:time_cost}
\end{equation}
where $d(u,v)$ is the Euclidean distance between nodes and $v_{\text{terrain}}(u,v)$ is the terrain- and slope-adjusted velocity extracted from the mobility matrix. In open, flat terrain, higher speeds can be achieved, leading to lower traversal times; conversely, restricted mobility conditions produce significantly higher costs, shaping the feasible search space and influencing both mission duration and exposure to threats.

\subsection{Optimization Cost (Log-Risk)}
The second metric, the optimization cost, quantifies the mission risk to be minimized. In the operational context considered, this risk corresponds to the likelihood of the \gls{ugv} being detected, targeted, or neutralized by enemy assets while executing its route. A per-node risk value $R(v) \in [0, 1]$ is pre-computed for every $v \in V$. This value is derived from a battlefield risk assessment framework that integrates multiple intelligence layers—such as known enemy positions, sensor ranges, and terrain masking—to estimate a survivability score for each location. For the interested reader, the model used to calculate these risk values, combining detection probability and aggregated impact, is described in detail in~\cite{soares2025argusframeworkriskawarepath}.

Formally, the total survival probability of a path $\pi$ is given by the product of the local survival probabilities across its vertices:
\begin{equation}
P_S(\pi) = \prod_{v \in \pi \setminus \{s\}} \bigl(1 - R(v)\bigr)
\label{eq:multiplicative_survival}
\end{equation}
Since shortest-path algorithms, including \gls{rcspp} solvers, require additive cost metrics, a logarithmic transformation is applied \cite{Mirsky2015, Cominetti2013}. The optimization cost $\ell(v)$ for each node is thus defined as its log-risk:
\begin{equation}
\ell(v) = -\log\bigl(1 - R(v)\bigr)
\label{eq:log_risk}
\end{equation}
Minimizing the cumulative log-risk along a path, $\sum_{v \in \pi} \ell(v)$, is mathematically equivalent to maximizing the total survival probability $P_S(\pi)$. This formulation enables the APULSE algorithm to efficiently identify the least risky feasible route, balancing traversal time and exposure in realistic military navigation scenarios.

\subsection{The RCSPP Mathematical Model}
\label{sec:rcspp_model}
Given a directed graph $G=(V,E)$, the additive resource cost $t(u,v)$ for each edge, and the additive optimization cost $\ell(v)$ for each node, the problem is formulated as a \gls{rcspp}.  

Given a start node $s$, a goal node $g$, and a maximum time budget $B$, the objective is to find a simple path $\pi$ from the set of all feasible paths $\mathcal{P}(s,g)$ that minimizes the cumulative log-risk while respecting the temporal constraint:
\begin{equation}
\min_{\pi\in\mathcal{P}(s,g)} \sum_{v\in\pi\setminus\{s\}} \ell(v)
\quad
\text{s.t.}\quad
 \sum_{(u,v)\in\pi} t(u,v) \leq B
\label{eq:rcspp_formulation}
\end{equation}

This formulation captures operational scenarios in which minimizing risk alone is insufficient—the mission must also be completed within a strict temporal limit. In tactical scenarios, such as coordinated maneuvers or time-sensitive reconnaissance, exceeding the time budget $B$ may compromise synchronization with other units or increase exposure to dynamic threats. The resulting path therefore represents the safest possible route that still satisfies the imposed time constraint, reflecting the optimal balance between survivability and timely arrival at the destination.

\section{The APULSE Algorithm}
\label{sec:apulse_algorithm}
To efficiently solve the \gls{rcspp} formulated in Section~\ref{sec:problem_formulation}, a hybrid label-setting algorithm named APULSE was developed, designed to achieve high performance on large-scale, grid-based graphs. The name APULSE reflects its synergistic nature, combining the goal-directed exploration of A* with the aggressive search-space reduction techniques of the Pulse family of algorithms~\cite{Cabrera2020}. This integration enables APULSE to inherit the heuristic efficiency of A* while leveraging Pulse-style pruning to discard unpromising paths early, substantially reducing computational overhead. As a result, APULSE is particularly effective in time-critical path-planning applications such as autonomous navigation over dense terrain graphs.

\subsection{Algorithmic Overview}
The APULSE algorithm is organized into two distinct phases that together ensure both optimality and scalability. The first is a pre-computation phase responsible for generating heuristic information that bounds the search and guides subsequent exploration. The second is the main label-setting loop, which iteratively expands partial paths using an informed best-first strategy, applying successive pruning checks—feasibility, optimality, and dominance—to remove suboptimal states before they grow the search space unnecessarily. This hierarchical pruning mechanism is crucial to control the combinatorial explosion typical of large \gls{rcspp} instances, allowing the algorithm to retain only the most promising candidates.

\begin{algorithm}[h!]
\caption{APULSE Algorithm}
\label{alg:apulse}
\begin{algorithmic}[1]
\STATE \textbf{Input:} Graph $G=(V,E)$, start $s$, goal $g$, budget $B$, target buckets $N$
\STATE \textbf{Output:} Optimal path $\pi^\star$
\STATE
\STATE \textbf{procedure} APULSE($G, s, g, B, N$)
\STATE $h_t \leftarrow$ \textproc{DijkstraReverse}($G, g$, "time")
\label{line:h_t}
\STATE $h_\ell \leftarrow$ \textproc{DijkstraReverse}($G, g$, "log-risk")
\label{line:h_ell}
\STATE
\STATE $\Delta T \leftarrow \max(1.0, B / N)$
\label{line:autotune}
\STATE
\STATE $Q \leftarrow$ \textproc{PriorityQueue}()
\STATE \textproc{push}($Q, (h_\ell[s], 0, 0, s, [s])$)
\STATE $visited \leftarrow \emptyset$
\STATE $g_\ell^\star \leftarrow \infty$; $\pi^\star \leftarrow \emptyset$
\STATE
\STATE \textbf{while} $Q$ is not empty \textbf{do}
\STATE \quad $(f, g_\ell, g_t, v, \pi) \leftarrow$ \textproc{pop}($Q$)
\STATE
\STATE \quad \textbf{if} $g_t + h_t[v] > B$ \textbf{then} \textbf{continue}
\label{line:prune_feasibility}
\STATE \quad \textbf{if} $f \ge g_\ell^\star$ \textbf{then} \textbf{continue}
\label{line:prune_optimality}
\STATE
\STATE \quad $b \leftarrow \lfloor g_t / \Delta T \rfloor$
\label{line:prune_bucket}
\STATE \quad \textbf{if} $(v, b) \in visited$ \textbf{and} $g_\ell \ge visited[(v, b)]$ \textbf{then}
\STATE \quad \quad \textbf{continue}
\STATE \quad $visited[(v, b)] \leftarrow g_\ell$
\STATE
\STATE \quad \textbf{if} $v = g$ \textbf{then}
\STATE \quad \quad $g_\ell^\star \leftarrow g_\ell$; $\pi^\star \leftarrow \pi$
\STATE \quad \quad \textbf{continue}
\STATE
\STATE \quad \textbf{for} $u \in \text{neighbors}(v)$ \textbf{do}
\STATE \quad \quad $g_t' \leftarrow g_t + t(v, u)$
\STATE \quad \quad $g_\ell' \leftarrow g_\ell + \ell(u)$
\STATE \quad \quad $f' \leftarrow g_\ell' + h_\ell[u]$
\STATE \quad \quad \textproc{push}($Q, (f', g_\ell', g_t', u, \pi + [u])$)
\STATE \textbf{end while}
\STATE
\STATE \textbf{return} $\pi^\star$
\end{algorithmic}
\end{algorithm}

\subsection{Pseudocode}
Algorithm~\ref{alg:apulse} summarizes the logic of APULSE, integrating the heuristic pre-computation, the A*-guided exploration, and the three-layer pruning mechanism into a single, efficient procedure. The pseudocode highlights the interplay between the algorithm’s components: the initialization of admissible heuristics, the dynamic bucketing of time states, and the label-expansion loop governed by a priority queue that enforces best-first ordering.

\subsection{Phases of the Algorithm}
The operation of APULSE is organized into two sequential phases. 
The first phase performs heuristic pre-computation to provide informed estimates for the subsequent search, while the second executes a guided exploration with multi-level pruning to ensure computational tractability. Together, these stages combine predictive insight with algorithmic efficiency, forming a cohesive framework for solving large-scale constrained path problems.

\subsubsection{Phase~1: Heuristic Pre-computation (lines~\ref{line:h_t}--\ref{line:h_ell})}
Before the main search is initiated, APULSE performs two reverse Dijkstra traversals starting from the goal node $g$. These preliminary runs compute:  
(1) the minimum-time heuristic $h_t(v)$, providing an admissible lower bound on the travel time from node $v$ to the goal; and  
(2) the minimum-risk heuristic $h_\ell(v)$, representing the minimum cumulative log-risk achievable from $v$ to $g$.  

Both heuristics are admissible and monotonic, ensuring the correctness of the A*-based exploration. Their joint use allows the algorithm to guide its expansion intelligently, focusing effort on the most promising regions of the search space and significantly reducing unnecessary exploration.

\subsubsection{Phase~2: Guided Search and Multi-Stage Pruning (lines~\ref{line:prune_feasibility}--\ref{line:prune_bucket})}
The main loop operates as a label-setting process ordered by the A*-style evaluation function $f(v) = g_\ell(v) + h_\ell(v)$, which balances accumulated and estimated log-risk. Each label generated during the expansion undergoes a three-stage pruning sequence designed to limit the combinatorial growth of candidate states:

\begin{enumerate}
    \item \textbf{Feasibility pruning} (line~\ref{line:prune_feasibility}) eliminates labels whose cumulative time $g_t(v)$ already exceeds the global time budget $B$, ensuring that only feasible paths remain.
    \item \textbf{Optimality pruning} (line~\ref{line:prune_optimality}) discards any label whose estimated total log-risk is no better than the best solution found so far, guaranteeing that the search remains competitive.
    \item \textbf{Dominance pruning via time bucketing} (line~\ref{line:prune_bucket}) addresses the exponential growth of non-dominated labels by discretizing the temporal dimension into fixed-width intervals (or “buckets”) of duration $\Delta T$. Each state is uniquely defined by the pair $(v, b)$, where $b = \lfloor g_t(v)/\Delta T \rfloor$. For each state, only the label with the lowest log-risk is retained; all others are pruned.
\end{enumerate}

This final mechanism—time bucketing—constitutes the cornerstone of APULSE’s scalability. By quantizing time, the algorithm trades a small, controllable loss in precision for orders-of-magnitude gains in computational efficiency.

\begin{figure}[t]
    \centering
    \includegraphics[width=0.9\columnwidth]{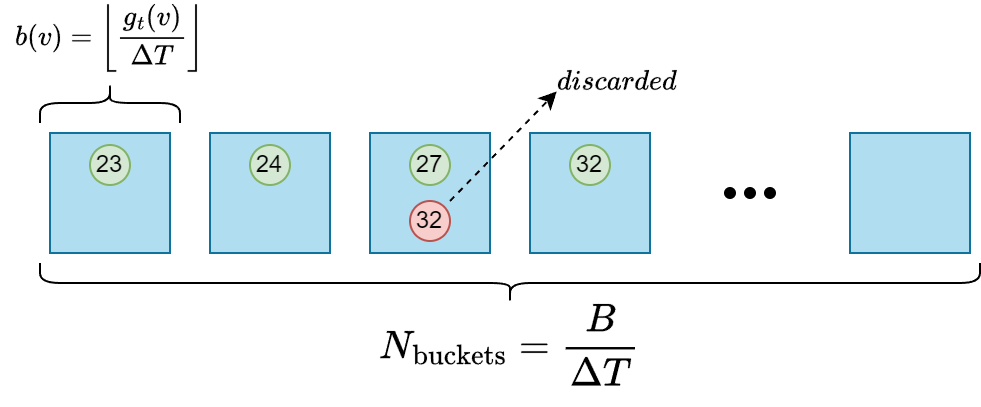}
    \caption{Illustration of the time bucketing mechanism. Multiple labels arriving at node $v$ are grouped into discrete time buckets. For each state $(v,b)$, only the label with the minimum log-risk (green dot) is retained, while all others (red dots) are pruned.}
    \label{fig:time_buckets}
\end{figure}

\subsubsection{Auto-Tuning of Temporal Granularity}
The bucket width $\Delta T$ introduces a tunable trade-off between precision and performance. A smaller $\Delta T$ increases temporal granularity, improving accuracy but generating more states; conversely, a larger $\Delta T$ reduces granularity and computation time at the cost of coarser solutions. Manual tuning of this parameter would be impractical, as the optimal value depends on the scale of each mission.

To address this, APULSE integrates an \textbf{auto-tuning heuristic} that adapts $\Delta T$ dynamically to the mission’s total time budget $B$. The heuristic sets $\Delta T$ according to a target number of temporal partitions $N$, as expressed in Eq.~\eqref{eq:bucket-auto-tuning}:
\begin{equation}
\Delta T \approx \frac{B}{N}
\label{eq:bucket-auto-tuning}
\end{equation}
This ensures that long-duration missions are represented with coarser time buckets, while shorter missions retain fine-grained temporal resolution. Empirically, a value of $N = 8192$ provides a robust balance between precision and runtime efficiency, preserving near-optimal solution quality across a wide range of problem scales without incurring excessive computational cost.

Together, these two phases—heuristic foresight and structured pruning—define an integrated framework that enables APULSE to achieve near-optimal solutions with a fraction of the computational effort required by conventional multi-objective search algorithms.

\section{Evaluation Results}
\label{sec:computational_study}
To evaluate how the APULSE algorithm performs against state-of-the-art \gls{rcspp} algorithms, a computational study was conducted. This benchmark, detailed in the following sections, was designed to compare the algorithms across several problem instances of increasing scale and complexity, all derived from a large, dense graph representative of complex planning problems. The evaluation focused on two key metrics: computational runtime, to assess efficiency and scalability, and solution quality, to ensure that performance gains did not come at the cost of significant sub-optimality.

\subsection{Experimental Setup}
\subsubsection{Hardware and Software}
All experiments were executed on a laptop equipped with an Intel\textsuperscript{\textregistered} Core\texttrademark{} Ultra 9 processor and 32 GB of RAM, running Ubuntu 22.04. The proposed APULSE algorithm was implemented in C++. Both APULSE and the competing algorithms, executed from the publicly available C++ repository of Ahmadi et al. \cite{Ahmadi2024EnhWCSP}, were compiled under identical conditions using the GNU Compiler Collection (GCC) version 11.4 to ensure a fair performance comparison. Each configuration was executed 5 times, with average runtimes recorded. Any run exceeding a ten-minute timeout was marked as a failure.

\subsubsection{Testbed Graph}
\begin{figure}[t]
    \centering
    \includegraphics[width=\columnwidth]{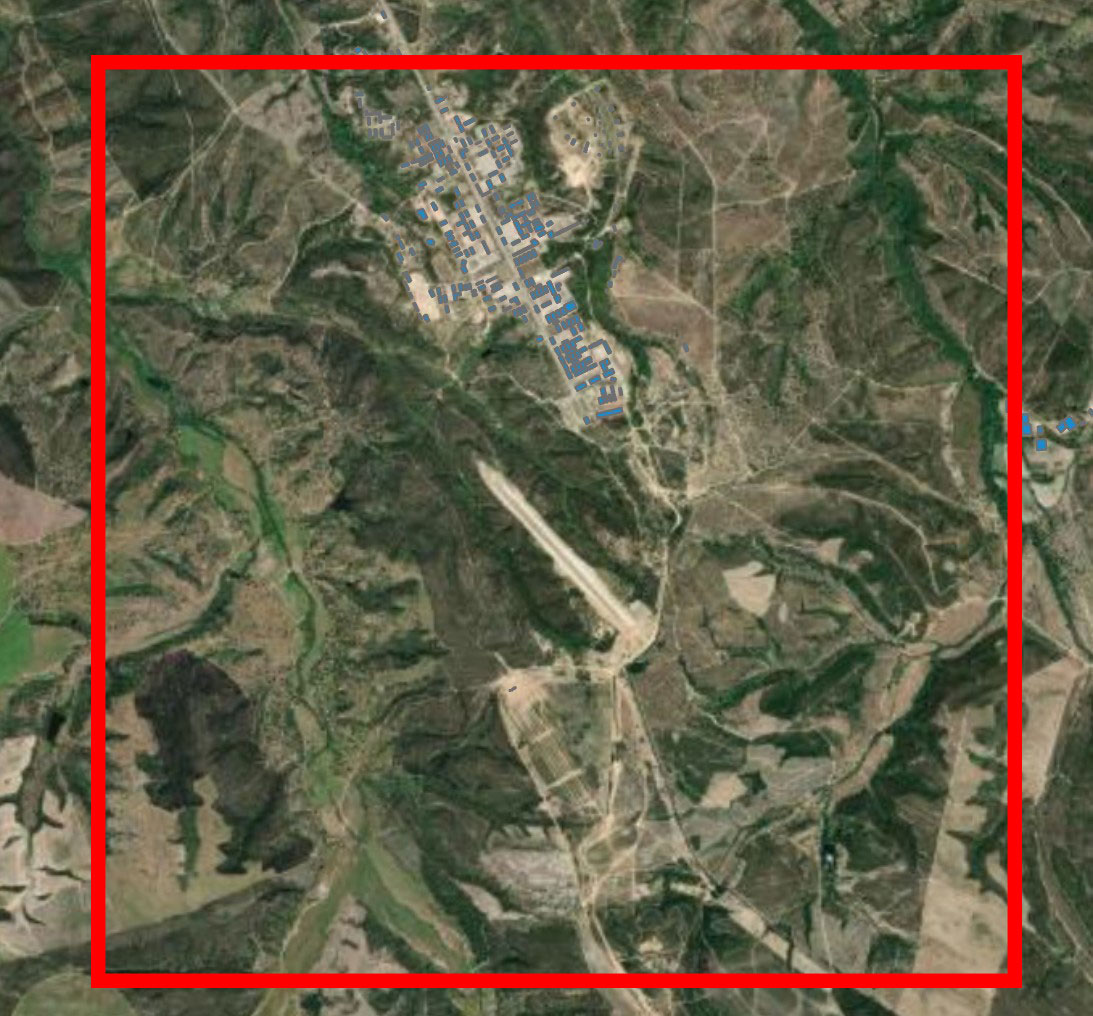}
    \caption{Operational test area (30~km\textsuperscript{2}) within the Santa Margarida military training field. The red boundary denotes the region used to construct the experimental graph.}
    \label{fig:graphic_map}
\end{figure}

The testbed for all experiments is a directed graph derived from a real-world, 30~km\textsuperscript{2} area of the Santa Margarida military training field in Portugal. The graph was constructed from geospatial data with a grid resolution of 25~meters, resulting in a large-scale network of 46,655 vertices and 361,276 edges. The operational area used in the experiments corresponds to the region circumscribed by the red boundary in Fig.~\ref{fig:graphic_map}.

For the purposes of mobility modelling and cost estimation, the terrain was categorised into six land-cover groups representing the dominant surface conditions in this region: \emph{Paved Areas}, \emph{Urban Areas}, \emph{Open Fields}, \emph{Light Vegetation}, \emph{Dense Scrub}, and \emph{Forest}. These classes directly influence the UGV's traversability profile, since each is associated with different mobility constraints and distinct entries in the terrain-dependent velocity matrix.

The elevation model for the same area ranges from a minimum altitude of 47~m to a maximum of 179~m, indicating moderate topographic variation. This altitude interval affects the slope-derived mobility penalties applied in the computation of terrain-adjusted velocities, which in turn modify the traversal time associated with each edge. Consequently, both the land-cover classes and the slope-induced speed reductions directly shaped the edge-cost landscape of the graph, influencing the behaviour of the search algorithms evaluated in the benchmark and contributing to the performance differences observed during experimentation.

\subsubsection{Problem Instances and Constraint Generation}
To evaluate performance across different problem scales, several test instances were defined by selecting different start-goal node pairs (1$\to$1k, 1$\to$15k, 1k$\to$25k, and 5k$\to$46k). To pinpoint the performance crossover, two additional instances (1$\to$5000 and 1$\to$7500) were also evaluated.

For each instance, the time budget \(B\) was systematically varied. First, the minimum possible travel time, \(T_{\text{min}}\), was determined. The budget \(B\) was then defined using a slack parameter \(\alpha \in [0.10, 1.00]\), where $B = T_{\min} \cdot (1 + \alpha)$. This allows testing the algorithms under different levels of constraint tightness.

\subsubsection{Competing Algorithms}
APULSE was compared against three reference competitors from the public repository of Ahmadi~\textit{et~al.}~\cite{Ahmadi2024EnhWCSP}: \gls{wc-a}, \gls{wc-ba}, and \gls{wc-ebba}. These algorithms represent the state-of-the-art in exact solvers for the \gls{rcspp} and provide a robust baseline for comparison.

\subsection{Performance Evaluation}
\label{sec:eval-benchmark-results}
The quantitative outcomes of the benchmarking campaign are presented across the following sections, analyzing runtime, scalability, and robustness.

\subsubsection{Runtime Under Tight and Moderate Constraints}
Table~\ref{tab:rcspp-main} summarises the initial results for tight to moderate budgets \(\alpha \in \{0.10, 0.20, 0.50\}\). All runtimes are reported in seconds, with failures or runs exceeding the ten-minute timeout denoted by “---”.

Across the main budgets, APULSE rapidly converges. At small scales, the overhead of the hybrid search is visible; from the medium scale onward, the trend reverses and APULSE consistently achieves the lowest average runtime. The differences become pronounced as the graph grows, with bidirectional methods increasingly affected by dominance-checking overheads.

\begin{table}[h!]
\centering
\caption{Average runtime (s) across scales and main budgets. Best performance is in \textbf{bold}. '---' indicates a timeout.}
\label{tab:rcspp-main}
\small
\sisetup{round-mode=places,round-precision=3,table-format=3.3}
\setlength{\tabcolsep}{6pt}
\resizebox{\columnwidth}{!}{%
\begin{tabular}{ll SSS}
\toprule
\textbf{Scale (start$\to$goal)} & \textbf{Algorithm} & {\(\alpha=0.10\)} & {\(\alpha=0.20\)} & {\(\alpha=0.50\)} \\
\midrule
\multirow{4}{*}{Small (1$\to$1k)}
& APULSE & 0.051 & 0.117 & 0.516 \\
& \gls{wc-a}& \bfseries 0.001 & \bfseries 0.001 & \bfseries 0.000 \\
& \gls{wc-ba}& 0.002 & 0.002 & 0.002 \\
& \gls{wc-ebba} & 0.001 & 0.001 & 0.001 \\
\midrule
\multirow{4}{*}{Medium (1$\to$15k)}
& APULSE & \bfseries 1.404 & \bfseries 5.327 & 27.005 \\
& \gls{wc-a}& 49.931 & 49.415 & 50.250 \\
& \gls{wc-ba}& 5.567 & 5.847 & \bfseries 11.790 \\
& \gls{wc-ebba} & 13.085 & 14.757 & 15.693 \\
\midrule
\multirow{4}{*}{Med-Large (1k$\to$25k)}
& APULSE & \bfseries 1.538 & \bfseries 4.737 & \bfseries 21.060 \\
& \gls{wc-a}& 27.617 & 29.206 & 30.302 \\
& \gls{wc-ba}& 103.567 & 105.380 & 97.331 \\
& \gls{wc-ebba} & 123.548 & 119.190 & 122.836 \\
\midrule
\multirow{4}{*}{Large (5k$\to$46k)}
& APULSE & \bfseries 1.191 & \bfseries 5.873 & \bfseries 21.602 \\
& \gls{wc-a}& 60.047 & 45.447 & 49.818 \\
& \gls{wc-ba}& {---} & {---} & {---} \\
& \gls{wc-ebba} & {---} & {---} & {---} \\
\bottomrule
\end{tabular}
}
\end{table}

\subsubsection{Impact of Constraint Tightness (High-Slack)}
To understand performance in less constrained scenarios, the budget slack \(\alpha\) was progressively increased. The results for these high-slack regimes, presented in Table~\ref{tab:rcspp-highslack}, highlight the conditions under which the advantage of APULSE over its competitors begins to diminish. As $\alpha$ increases, the runtime of APULSE grows moderately, and its advantage over \gls{wc-a} gradually diminishes. This trend culminates at the "Large" scale for $\alpha = 1.0$, where both algorithms exhibit comparable performance, while the bidirectional variants continue to fail.

\begin{table}[h!]
\centering
\caption{Average runtime (s) in high-slack regimes ($\alpha > 0.5$).}
\label{tab:rcspp-highslack}
\small
\sisetup{round-mode=places,round-precision=3,table-format=3.3}
\setlength{\tabcolsep}{6pt}
\resizebox{\columnwidth}{!}{%
\begin{tabular}{ll SSS}
\toprule
\textbf{Scale (start$\to$goal)} & \textbf{Algorithm} & {\(\alpha=0.60\)} & {\(\alpha=0.80\)} & {\(\alpha=1.00\)} \\
\midrule
\multirow{4}{*}{Medium (1$\to$15k)}
& APULSE & 31.686 & 31.771 & 34.960 \\
& \gls{wc-a}& 54.401 & 46.631 & 48.090 \\
& \gls{wc-ba}& 16.325 & \bfseries 12.739 & \bfseries 12.800 \\
& \gls{wc-ebba} & \bfseries 15.191 & 13.431 & 12.842 \\
\midrule
\multirow{4}{*}{Med-Large (1k$\to$25k)}
& APULSE & \bfseries 15.369 & \bfseries 16.319 & \bfseries 16.845 \\
& \gls{wc-a}& 28.143 & 26.986 & 31.635 \\
& \gls{wc-ba}& 96.136 & 92.062 & 113.824 \\
& \gls{wc-ebba} & 105.295 & 122.217 & 127.645 \\
\midrule
\multirow{4}{*}{Large (5k$\to$46k)}
& APULSE & \bfseries 33.921 & \bfseries 43.804 & 50.295 \\
& \gls{wc-a}& 59.353 & 49.585 & \bfseries 50.262 \\
& \gls{wc-ba}& {---} & {---} & {---} \\
& \gls{wc-ebba} & {---} & {---} & {---} \\
\bottomrule
\end{tabular}
}
\end{table}

\subsubsection{Performance Crossover Analysis}
To estimate the instance size at which APULSE begins to gain a consistent advantage, the two additional test configurations (1$\to$5000 and 1$\to$7500) were introduced. Their results are summarised in Table~\ref{tab:rcspp-crossover}.

\begin{table}[h!]
\centering
\caption{Average runtime (s) for crossover analysis instances. 'n/a' indicates tests not performed.}
\label{tab:rcspp-crossover}
\small
\setlength{\tabcolsep}{5pt}
\sisetup{round-mode=places,round-precision=3,table-format=3.3}
\resizebox{\columnwidth}{!}{%
\begin{tabular}{ll S S S}
\toprule
\textbf{Instance} & \textbf{Algorithm} & {\(\alpha=0.10\)} & {\(\alpha=0.20\)} & {\(\alpha=0.50\)} \\
\midrule
\multirow{4}{*}{1$\to$1000}
& APULSE & 0.042 & 0.124 & 0.481 \\
& \gls{wc-a}& \bfseries 0.001 & \bfseries 0.001 & \bfseries 0.001 \\
& \gls{wc-ba}& 0.002 & 0.002 & 0.002 \\
& \gls{wc-ebba} & 0.001 & 0.001 & 0.001 \\
\midrule
\multirow{4}{*}{1$\to$5000}
& APULSE & \bfseries 0.770 & \bfseries 0.252 & 0.030 \\
& \gls{wc-a}& 52.182 & 46.228 & \bfseries 0.004 \\
& \gls{wc-ba}& 10.390 & 93.950 & 9.526 \\
& \gls{wc-ebba} & 9.985 & 11.045 & 10.887 \\
\midrule
\multirow{4}{*}{1$\to$7500}
& APULSE & \bfseries 1.589 & \bfseries 6.885 & {n/a} \\
& \gls{wc-a}& 435.610 & 434.116 & {n/a} \\
& \gls{wc-ba}& 326.254 & 324.959 & {n/a} \\
& \gls{wc-ebba} & 378.912 & 390.415 & {n/a} \\
\bottomrule
\end{tabular}
}
\end{table}

\begin{figure}[h!]
\centering
\includegraphics[width=\columnwidth]{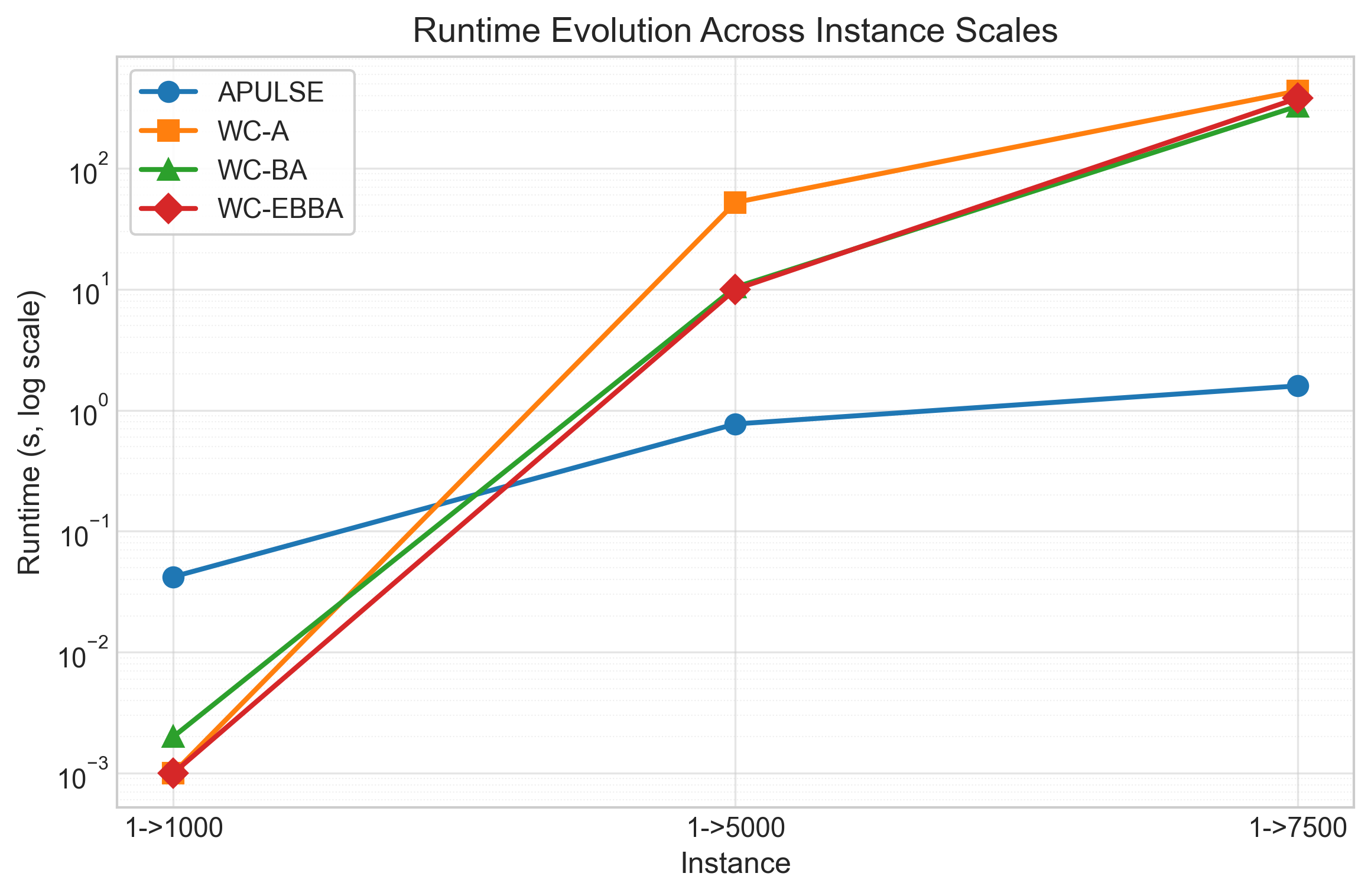}
\caption{Runtime evolution across instance scales ($\alpha=0.10$). APULSE becomes consistently faster from the 1$\to$5000 configuration onwards, confirming the crossover point.}
\label{fig:crossover}
\end{figure}

Fig.~\ref{fig:crossover} illustrates the runtime evolution on a logarithmic scale for a tight budget (\(\alpha=0.10\)). The reference algorithms initially perform faster on the smallest instance, but their runtimes increase sharply with problem size. By the 1$\to$7500 instance, all competitors are several orders of magnitude slower than APULSE. This pattern reflects the exponential growth of the feasible space as the graph becomes denser with near-optimal paths, requiring more exhaustive exploration. Under these conditions, APULSE’s hybrid design demonstrates superior scalability, with the performance crossover occurring at approximately 3000 nodes.

\begin{figure}[t]
\centering
\subfloat[]{\includegraphics[width=\columnwidth]{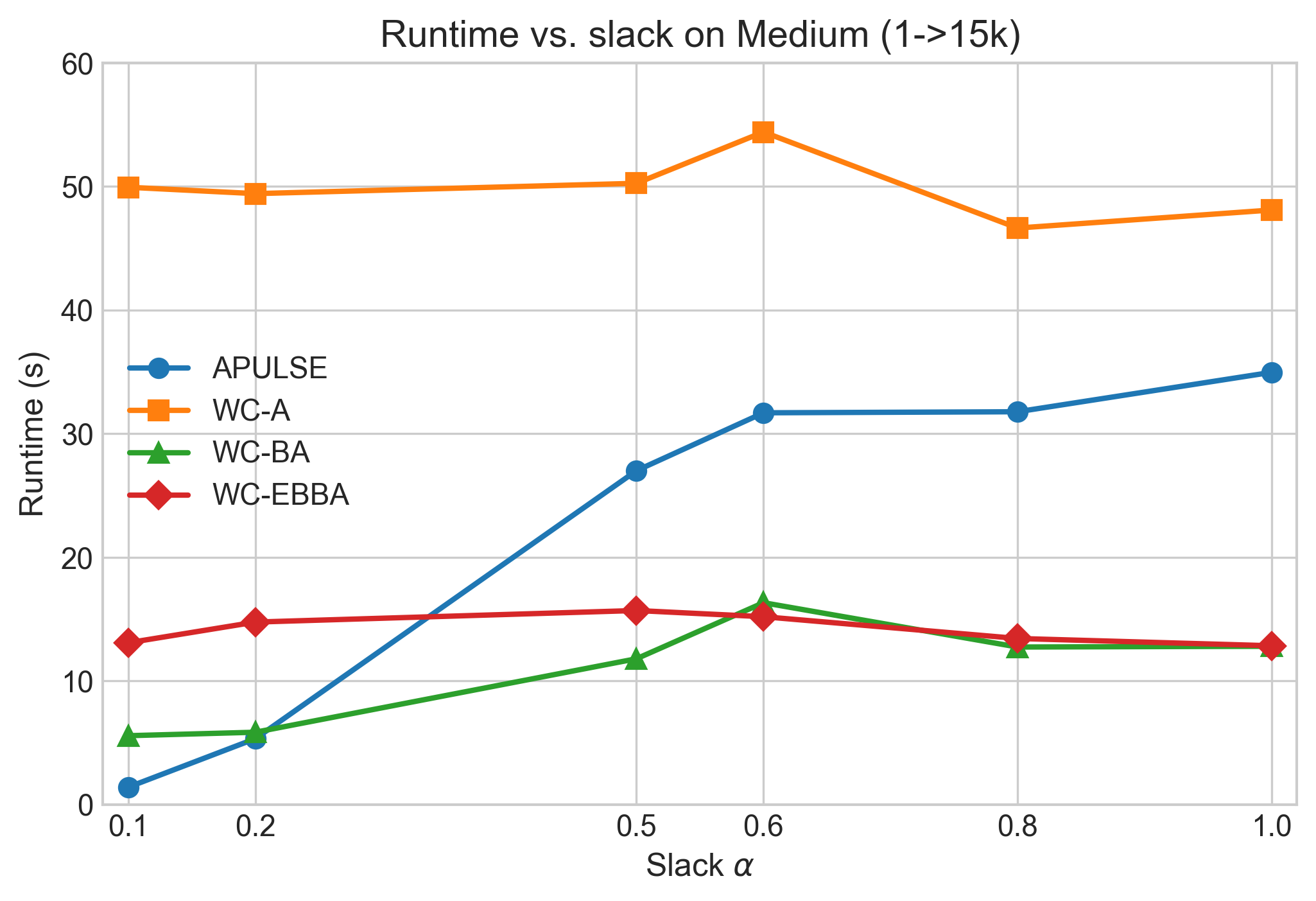}}
\\ 
\vspace{5pt} 
\subfloat[]{\includegraphics[width=\columnwidth]{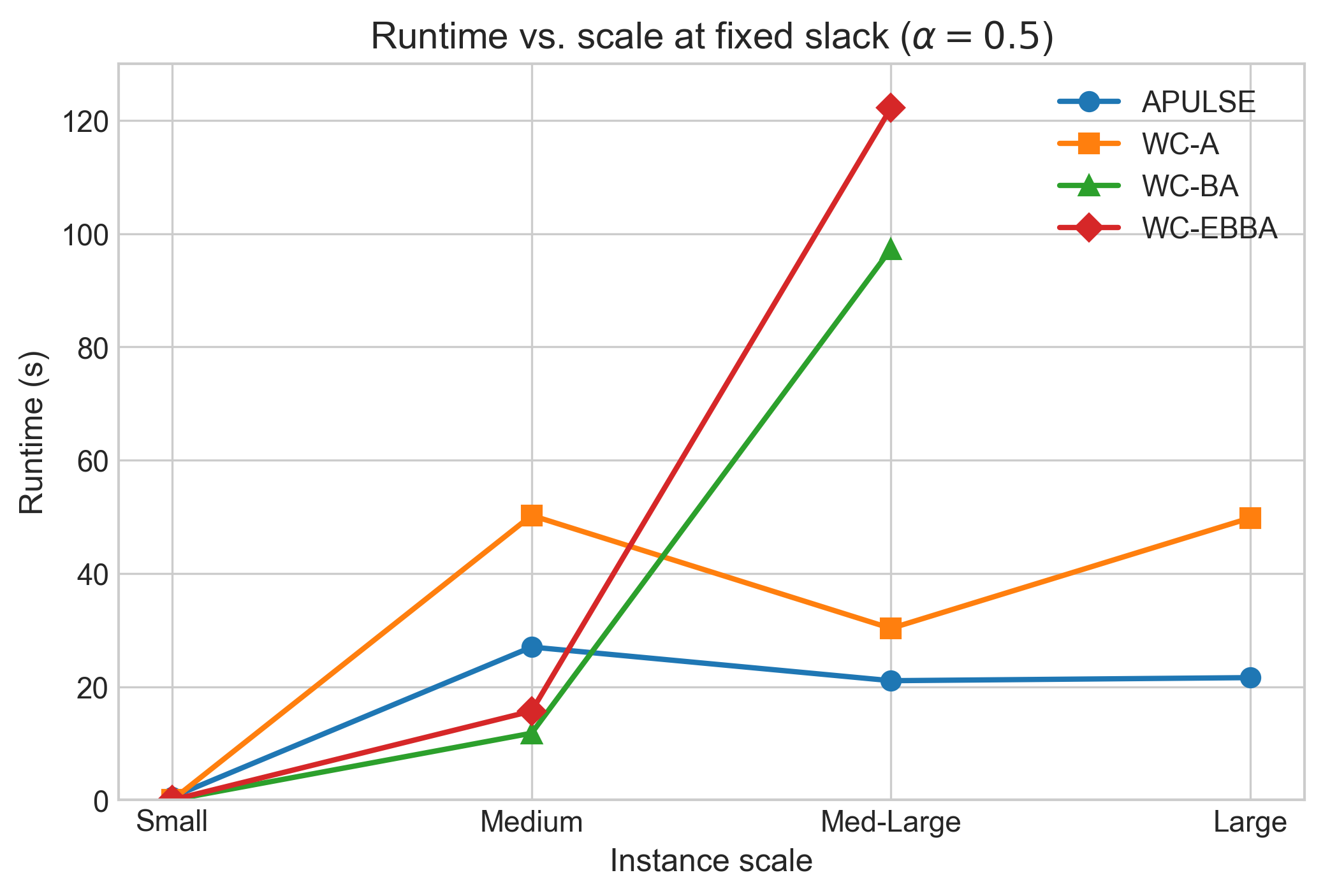}}

\caption{Runtime performance of APULSE compared to reference algorithms under two factors. (a) Effect of increasing budget slack $\alpha$ on runtime (Medium instance, 1$\to$15k). (b) Runtime scaling across instance sizes for a fixed slack $\alpha=0.5$.}
\label{fig:bench_results_stack}
\end{figure}

The behaviour of the algorithms was analysed with respect to the two main factors that influence runtime: the slack parameter $\alpha$ and the instance scale. 
Fig.~\ref{fig:bench_results_stack}(a) reports runtime as a function of slack $\alpha$ for the Medium instance (1$\to$15k), 
while Fig.~\ref{fig:bench_results_stack}(b) shows how runtime scales with instance size for a fixed slack of $\alpha=0.5$.

\subsection{Solution Quality Analysis}
\label{sec:eval-quality}
While the preceding sections focused on runtime, solution quality was validated by comparing the returned path ($\rho_{\text{APULSE}}$) against the known exact optimal risk path ($\rho^*$) across all $26$ test configurations.

The results, summarized in Table~\ref{tab:rcspp-quality}, confirm that APULSE consistently achieved the optimal solution ($\rho_{\text{APULSE}} = \rho^*$) in 25 out of 26 instances. The single suboptimal result occurred at the Med-Large scale ($1\text{k}\to 25\text{k}$) with $\alpha = 0.50$, where the risk deviated from the optimal by only 0.0025\%. This negligible loss of optimality is an expected consequence of the time-bucketing mechanism. This outcome validates the design choice discussed in Section \ref{sec:apulse_algorithm}, demonstrating that the auto-tuning heuristic achieves superior computational performance without meaningfully compromising solution quality.

\begin{table}[t]
\centering
\caption{Optimality of the APULSE solution compared to the exact optimal path ($\rho^*$).}
\label{tab:rcspp-quality}
\setlength{\tabcolsep}{14pt}
\renewcommand{\arraystretch}{1.2}
\normalsize
\begin{tabular}{lc}
\toprule
\textbf{Optimality Outcome} & \textbf{Instances ($N=26$)} \\
\midrule
Optimal Path Achieved & 25 \\
Suboptimal Path Found & 1 \\
\bottomrule
\end{tabular}
\end{table}

\section{Discussion}
\label{sec:discussion}

The computational study in Section~\ref{sec:computational_study} provides empirical evidence on the performance of APULSE for solving the \gls{rcspp} on dense, grid-based graphs. The results highlight clear differences in scalability across the evaluated algorithms. While the state-of-the-art algorithm \gls{wc-a} is highly efficient on the smallest ``Small'' instances as shown in Table~\ref{tab:rcspp-main}, its runtime increases substantially as problem size grows. In contrast, although APULSE is slower on the smallest instance, it exhibits more favourable scalability, consistently outperforming \gls{wc-a} on most ``Medium'' and ``Large'' instances across all tested slack parameters (as presented in Tables~\ref{tab:rcspp-main} and \ref{tab:rcspp-highslack}). This suggests that while \gls{wc-a} remains effective for sparse or small-scale scenarios, its performance is less robust under the large, high-connectivity graphs used in this testbed.

A key observation is the consistent failure of the bidirectional algorithms \gls{wc-ba} and \gls{wc-ebba} on the larger instances, as shown in Table~\ref{tab:rcspp-main}. This behaviour suggests that these approaches are unsuitable for this problem domain, likely due to an explosion of non-dominated labels at both search frontiers when the graph density increases. APULSE's hybrid design—combining A* guidance with state-space reduction through time-bucketing—directly mitigates this effect. As illustrated in Fig.~\ref{fig:crossover}, whereas the competitors' runtimes grow exponentially, APULSE scales at a noticeably slower rate.

The solution-quality analysis Table~\ref{tab:rcspp-quality} quantifies the trade-off inherent to APULSE's design. The algorithm achieved the exact optimal solution in 25 of the 26 benchmark instances, with a deviation of only 0.0025\% in the single outlier. This confirms that the auto-tuning heuristic for bucket sizes Eq.~\ref{eq:bucket-auto-tuning} achieves a functional balance between computational efficiency and optimality. APULSE therefore provides high-quality solutions in timeframes where exact solvers may be computationally prohibitive.

A notable practical implication of this efficiency is that it shifts the \gls{rcspp} from a purely offline computation to a viable interactive decision-support capability. The sub-minute runtimes achieved by APULSE—even on the 46k-node ``Large'' instance—enable two operationally relevant features.

\begin{figure}[t]
    \centering
    \includegraphics[width=\columnwidth]{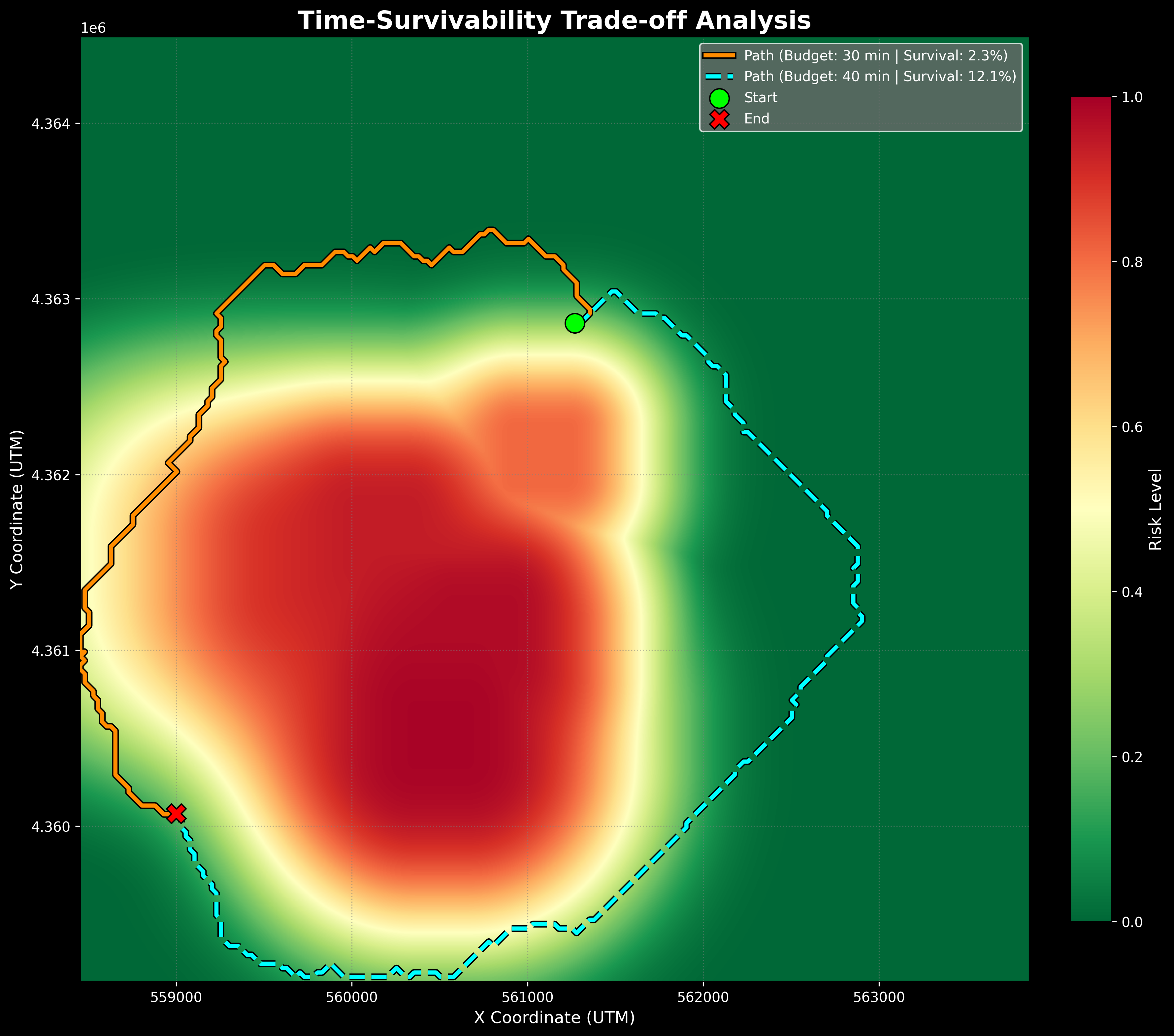}
    \caption{Time–survivability trade-off analysis. The orange route corresponds to a 30-minute budget (2.3\% survival), while the cyan route represents a 40-minute budget (12.1\% survival). Extending the time budget enables the algorithm to avoid the central high-risk region and identify a substantially safer path.}
    \label{fig:tradeoff_plot}
\end{figure}

First, APULSE enables \textbf{interactive trade-off analysis}: an operator can repeatedly solve the problem under different time budgets ($B$), thereby quantitatively exploring how mission duration influences survivability. As shown in Fig.~\ref{fig:tradeoff_plot}, a strict 30-minute budget results in a route with only 2.3\% survival probability, whereas relaxing the budget to 40 minutes yields a markedly safer alternative with 12.1\% survival. Increasing the allowed time by merely ten minutes (33\%) enables the algorithm to bypass the central high-risk corridor, an insight only possible with a high-speed solver capable of evaluating multiple alternative runs within the short decision windows imposed by the tactical environment.

Second, the algorithm’s performance enables \textbf{dynamic replanning}. Because APULSE can recompute the core problem within seconds, its optimised structure makes it particularly well suited for local route recalculations in real time. This includes reacting to newly detected threats, unexpected enemy movements, or the sudden appearance of obstacles. As a result, the planner can continuously update and refine the trajectory, maintaining an optimal or near-optimal path as the tactical situation evolves.

Finally, it is important to contextualize these findings. APULSE was validated on static, log-additive cost graphs that represent realistic yet controlled mission settings. The algorithmic framework presented here should therefore be viewed as a robust and scalable baseline for future extensions of constrained path-planning systems—particularly those incorporating dynamic environmental updates, probabilistic uncertainty, or coordinated multi-agent operations.

\section{Conclusion}
\label{sec:conclusion}
This paper introduced APULSE, a hybrid algorithm designed to efficiently solve the computationally challenging \gls{rcspp} on large-scale dense graphs. The core contribution is the empirical demonstration of its scalability and robustness in a problem domain—dense grids—where established state-of-the-art exact solvers (\gls{wc-ba}, \gls{wc-ebba}) fail to converge. Through a computational study, it was shown that APULSE finds near-optimal solutions, achieving exact optimality in 25 of 26 tested configurations, while running orders of magnitude faster than competing methods on larger instances.

The significance of this performance is that it demonstrates the viability of using sophisticated, optimization-based models in time-critical operational cycles. The sub-minute solution times provide a practical foundation for real-world decision-support tools that require rapid route generation and timely analysis of mission alternatives.

This work opens several avenues for future research. From an algorithmic perspective, the hybrid design of APULSE may be further strengthened by improving the time-bucketing strategy to balance pruning efficiency and final solution quality. From an application standpoint, the optimization framework can be expanded to incorporate additional operational constraints, such as modelling vehicle energy as a second resource, integrating non-holonomic kinematics, or accounting for communication link quality. Finally, the algorithmic foundation established here provides a scalable basis for addressing the more complex problem of multi-agent constrained path planning.

\section*{Acknowledgment}
This work was conducted within the scope of the Portuguese Army Project EXE03 -- Unmanned Ground Systems, which is part of the EXE02 -- Remote and Autonomous Systems portfolio. This work was also funded by national funds through FCT --- Fundação para a Ciência e a Tecnologia, I.P., under projects/supports UID/6486/2025 (\url{https://doi.org/10.54499/UID/06486/2025}) and UID/PRR/6486/2025 (\url{https://doi.org/10.54499/UID/PRR/06486/2025}).

\bibliographystyle{IEEEtran}
\bibliography{referencias}

@article{Lozano2013,
  author    = {Leonardo Lozano and A. Medaglia},
  title     = {On an exact algorithm for the constrained shortest path problem},
  journal   = {Computers \& Operations Research},
  volume    = {40},
  number    = {3},
  pages     = {853--862},
  year      = {2013},
  doi       = {10.1016/j.cor.2012.10.007}
}

@article{Pugliese2013,
  author    = {L. Pugliese and F. Guerriero},
  title     = {A survey of resource constrained shortest path problems: Exact solution approaches},
  journal   = {Networks},
  volume    = {62},
  number    = {3},
  pages     = {143--164},
  year      = {2013},
  doi       = {10.1002/net.21500}
}

@article{koenig2004lifelong,
  title     = {Lifelong planning A*},
  author    = {Koenig, Sven and Likhachev, Maxim and Furcy, David},
  journal   = {Artificial Intelligence},
  volume    = {155},
  number    = {1-2},
  pages     = {93--146},
  year      = {2004},
  publisher = {Elsevier}
}

@inproceedings{Ahmadi2021,
  author    = {Saman Ahmadi and P. Kilby and Abdul B. Sattar},
  title     = {A Fast Exact Algorithm for the Resource Constrained Shortest Path Problem},
  booktitle = {Proceedings of the International Conference on Automated Planning and Scheduling},
  year      = {2021},
  volume    = {31},
  pages     = {1--9},
  doi       = {10.1609/icaps.v31i1.15948}
}

@article{Thomas2018,
  author    = {Barrett W. Thomas and Mike Hewitt},
  title     = {An exact bidirectional A* approach for solving resource-constrained shortest path problems},
  journal   = {Networks},
  volume    = {72},
  number    = {1},
  pages     = {79--97},
  year      = {2018},
  doi       = {10.1002/net.21817}
}

@article{Hart1968,
  author    = {P. Hart and N. Nilsson and B. Raphael},
  title     = {A Formal Basis for the Heuristic Determination of Minimum Cost Paths},
  journal   = {IEEE Transactions on Systems Science and Cybernetics},
  volume    = {4},
  number    = {2},
  pages     = {100--107},
  year      = {1968},
  doi       = {10.1109/TSSC.1968.300136}
}

@inproceedings{Holte2016,
  author    = {Robert Holte and Ariel Felner and Guni Sharon and Nathan Sturtevant},
  title     = {Bidirectional Search That Is Guaranteed to Meet in the Middle},
  booktitle = {Proceedings of the AAAI Conference on Artificial Intelligence},
  volume    = {30},
  number    = {1},
  year      = {2016},
  doi       = {10.1609/aaai.v30i1.10436}
}

@article{Cabrera2020,
author = {Cabrera, Nicolás and Medaglia, Andrés L. and Lozano, Leonardo and Duque, Daniel},
title = {An exact bidirectional pulse algorithm for the constrained shortest path},
journal = {Networks},
volume = {76},
number = {2},
pages = {128-146},
doi = {https://doi.org/10.1002/net.21960},
year = {2020}
}

@incollection{Irnich2005,
  author    = {Irnich, Stefan and Desaulniers, Guy},
  title     = {Shortest Path Problems with Resource Constraints},
  booktitle = {Column Generation},
  publisher = {Springer US},
  year      = {2005},
  pages     = {33-65},
  editor    = {Desaulniers, Guy and Desrosiers, Jacques and Solomon, Marius M.},
  doi       = {10.1007/0-387-25486-2_2}
}

@incollection{Delling2009,
  author    = {Delling, Daniel and Sanders, Peter and Schultes, Dominik and Wagner, Dorothea},
  title     = {Engineering Route Planning Algorithms},
  booktitle = {Algorithmics of Large and Complex Networks},
  publisher = {Springer Berlin Heidelberg},
  year      = {2009},
  pages     = {117-139}
}

@article{Ahmadi2024EnhWCSP,
  author    = {Ahmadi, Saman and Tack, Guido and Harabor, Daniel and Kilby, Philip and Jalili, Mahdi},
  title     = {Enhanced methods for the weight constrained shortest path problem},
  journal   = {Networks},
  volume    = {84},
  number    = {1},
  pages     = {3-30},
  doi       = {https://doi.org/10.1002/net.22210},
  year      = {2024}
}

@misc{cos_dgt,
  author       = {{Direção-Geral do Território}},
  title        = {Carta de Uso e Ocupação do Solo (COS)},
  year         = {2025},
  howpublished = {\url{https://smos.dgterritorio.gov.pt/cartografia-de-uso-e-ocupacao-do-solo}},
  note         = {Acedido em 24 de setembro de 2025}
}

@misc{soares2025argusframeworkriskawarepath,
  title         = {ARGUS: A Framework for Risk-Aware Path Planning in Tactical UGV Operations}, 
  author        = {Nuno Soares and António Grilo},
  year          = {2025},
  eprint        = {2511.07565},
  archivePrefix = {arXiv},
  primaryClass  = {eess.SY},
  url           = {https://arxiv.org/abs/2511.07565}
}

@inproceedings{Mirsky2015,
  author    = {Yisroel Mirsky and Amiram Bar-Toal and Meir Kalech and Bracha Shapira},
  title     = {Search Problems in the Domain of Multiplication: Case Study on Anomaly Detection Using Markov Chains},
  booktitle = {Proceedings of the Twenty-Ninth AAAI Conference on Artificial Intelligence},
  year      = {2015},
  pages     = {1641--1647}
}

@article{Cominetti2013,
  author    = {R. Cominetti and Alfredo Torrico},
  title     = {Additive Consistency of Risk Measures and Its Application to Risk-Averse Routing in Networks},
  journal   = {Mathematics of Operations Research},
  volume    = {38},
  number    = {1},
  pages     = {44--60},
  year      = {2013},
  doi       = {10.1287/moor.1120.0559}
}


\begin{IEEEbiography}[{\includegraphics[width=1in,height=1.25in,clip,keepaspectratio]{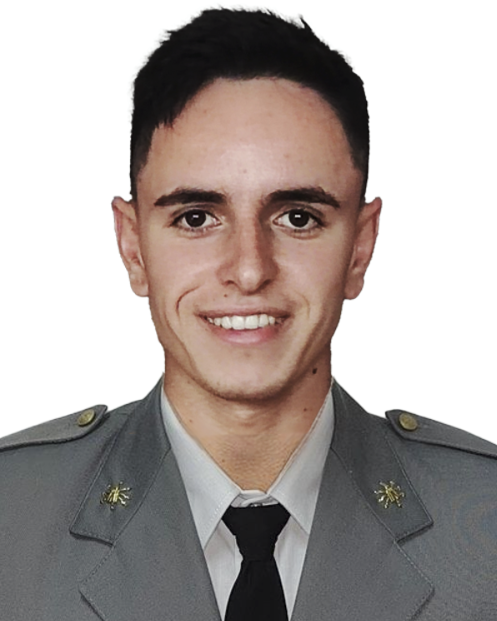}}]{Nuno Soares}
received the B.Sc.\ degree in Military Eletrical and Computers Engineering from the Portuguese Military Academy, Lisbon, Portugal, in 2023. He serves as an Officer in the Portuguese Army within the Signals branch. His professional and research interests include artificial intelligence models and optimization methods applied to autonomous systems and military decision-support processes.
\end{IEEEbiography}


\begin{IEEEbiography}[{\includegraphics[width=1in,height=1.25in,clip,keepaspectratio]{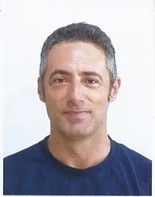}}]{Prof.\ António Grilo}
received the Ph.D.\ degree in Electrical and Computer Engineering from Instituto Superior Técnico (IST), University of Lisbon, where he is currently an Associate Professor. At IST, he teaches courses in Mobile Communications, Internet of Things (IoT), and Smart Grids. Since 1996, he has participated in multiple European Commission (EC) projects related to communication networks. He is presently a Researcher with the Intelligent Communication Networks Group at INESC INOV-Lab, Lisbon. He is the author or co-author of more than eighty scientific publications in the field of communication networks. His current research interests include UxV networks, the Internet of Things, edge computing optimization, and the application of artificial intelligence to communication systems and networks. He is a Senior Member of the IEEE and a Life Member of AFCEA.
\end{IEEEbiography}

\vfill



\end{document}